# Ahead of the Text: Leveraging Entity Preposition for Financial Relation Extraction

Stefan Pasch
Frankfurt, Germany
stefan.pasch@outlook.com

Dimitrios Petridis
Frankfurt, Germany
petridis.dim@outlook.com

## ABSTRACT

In the context of the ACM KDF-SIGIR 2023 competition, we undertook an entity relation task on a dataset of financial entity relations called REFind. Our top-performing solution involved a multi-step approach. Initially, we inserted the provided entities at their corresponding locations within the text. Subsequently, we fine-tuned the transformer-based language model roberta-large for text classification by utilizing a labeled training set to predict the entity relations. Lastly, we implemented a post-processing phase to identify and handle improbable predictions generated by the model. As a result of our methodology, we achieved the 1st place ranking on the competition's public leaderboard.

## CCS CONCEPTS

• Information Systems → Information Retrieval

## KEYWORDS

Financial Relation Extraction, NLP, transformer models, 10-X reports

**ACM Reference format:**

Stefan Pasch, Dimitrios Petridis 2023. Ahead of the Text: Leveraging Entity Preposition for Financial Relation Extraction. *ACM SIGIR: The 4th Workshop on Knowledge Discovery from Unstructured Data in Financial Services (SIGIR-KDF '23)*.

## 1 Introduction

This paper aims to briefly summarize and present our work and findings, related to the Relation Extraction Challenge, organized by KDF.SIGIR, and sponsored by J.P. Morgan. Relation Extraction is an NLP task of automatically identifying and classifying the semantic relationships that may exist between different entities in a given text. The challenge consisted of two phases, the development, and the final phase. It is solely based on the REFinD dataset [2], and this shared task is part of the Fourth Workshop on Knowledge Discovery from Unstructured Data in Financial Services. The REFinD dataset is the first domain-specific financial relation-extraction dataset built using raw text from various 10-X reports of publicly traded companies that were obtained from the US Securities and Exchange Commission (SEC) website. In total, it consists of over 30.000 instances and 22 types of relations amongst 8 types of entity pairs, generated entirely over financial documents.

The entities include organization, person, geographic location, date, money, university, and government-agency, and the entity pairs consist of meaningful matchings between two entities, like organization-date or person-university. Finally, the matching explains the relation between the entity pairs, for instance *org-date:formed_on* when the date indicates the foundation date of the company. For a given entity pair, the relation can also always be a "No Relation", in case there is no identified relation in line with the 22 pre-defined relation-groups. Table 1 shows all entity-pairs and relations with the overall number of labeled observations.

**TABLE 1: Number of Observations per Entity Pair and Relation**

| Entity Pair | Obs Ent Pair | Relation | Obs Relation |
|---|---|---|---|
| ORG-ORG | 6332 | org:org:agreement_with | 935 |
| | | org:org:subsidiary_of | 551 |
| | | org:org:shares_of | 408 |
| | | org:org:acquired_by | 78 |
| ORG-GPE | 6080 | org:gpe:operations_in | 4043 |
| | | org:gpe:headquartered_in | 193 |
| | | org:gpe:formed_in | 115 |
| PERSON-TITLE | 5471 | pers:title:title | 4468 |
| ORG-DATE | 4919 | org:date:formed_on | 640 |
| | | org:date:acquired_on | 186 |
| PERSON-ORG | 4007 | pers:org:employee_of | 2479 |
| | | pers:org:member_of | 630 |
| | | pers:org:founder_of | 131 |
| ORG-MONEY | 1519 | org:money:revenue_of | 311 |
| | | org:money:loss_of | 202 |
| | | org:money:profit_of | 29 |
| | | org:money:cost_of | 23 |
| PERSON-UNIV | 188 | pers:univ:employee_of | 76 |
| | | pers:univ:attended | 43 |
| | | pers:univ:member_of | 33 |
| PERSON-GOV_AGY | 160 | pers:gov_agy:member_of | 56 |
| | | no_relation | 13046 |

During the development phase of the challenge, we received three public datasets (train, dev, test) of 28.676 instances in total, which were labeled with the actual relations between the two already identified entities. Additionally, all three datasets were enriched with all intermediate variables that were generated during the named entity tagging process using the spaCy library, by the original authors [2]. During the first weeks of the development phase, we explored the public datasets, performed various tests and tabulations, and focused on understanding the actual conceptual relations within the texts, by manually investigating indicative data points based on the actual labels. Later, in our attempt to find the most appropriate model, we applied multiple transformers models, by testing them on the public dataset as a whole. In order to further boost our results, and provide a more accurate and specific input to the model, we incorporated the group entities, by inserting them inside the raw text. After testing multiple models and input combinations, we ended up leveraging Roberta-Large for text classification.

In the development phase, we trained the model using the train set of the public dataset, resulting in an F1-score of around 75%. Moving to the final phase, we received the private, unlabeled dataset of additional 3069 instances. Using our pre-trained model, and similarly enhancing the text with the group entities, we performed the predictions. This resulted in relatively high F1-scores (up to 72.6%), but we managed to further increase our accuracy in a post-processing step that involved various plausibility checks.

## 2 Methodology

Given the predefined 22 relation categories, we conducted the financial relation extraction challenge as a text classification task with the 22 relations as outcome groups. However, to capture the given entity pairs we modified the input text in the following way.

### 2.1 Text Pre-Processing

Similar to the matching-the-blanks approach [4], we inserted entity markers for e1 and e2 inside the text. However, we did not simply add general markers for entities, like "ent1" and "ent2", but instead included a marker specific to the corresponding entity, e.g. ORG for organization or PERS for person. We experimented with three different approaches to include these entity information inside the text:

i) Adding the entity marker before the entity: PERS John Doe is the CEO of ORG Company A.
ii) Adding the entity marker before and after the entity: PERS John Doe PERS is the CEO of ORG Company A ORG.
iii) Adding an entity-relation marker before the entire text: <PERS-ORG> John Doe is the CEO of Company A.

### 2.2 Text Classification

To classify these pre-processed texts, we fine-tuned transformer-based models for text classification. In particular, we fine-tuned roberta-large for the text classification task at hand [3]. After hyperparameter optimization during the development phase, we applied the hyperparameters as shown in Table 2. We also tested bert-base-uncased [1] but received considerably lower scores.

**TABLE 2: Hyperparameters for Text Classification Fine-Tuning**

| Hyperparameter | Value |
| --- | --- |
| Learning Rate | 1e-5 |
| Epochs | 3 |
| Batch Size | 16 |
| Weight Decay | 0.01 |
| Optimizer | Adam Optimizer |

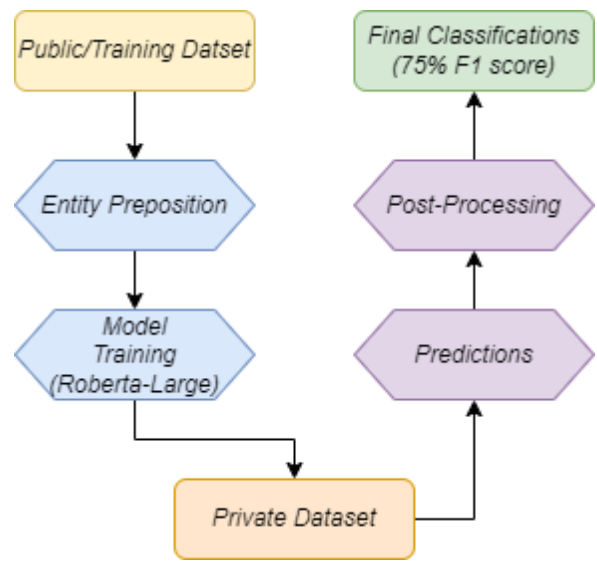

**Figure 1: Workflow Entity Relation Extraction**

### 2.3 Post-Processing

After receiving the model's prediction on the test set, we conducted a post-processing step on the outcomes. The nature of the dataset allows for a plausibility check regarding the entity pairs. For example, the label 6 "org:org:acquired_by" is only plausible if the initial entity pair were both of type ORG. If, however, the label 6 was predicted on a given entity pair of type ORG-DATE, we know that the prediction is wrong. Therefore, in cases where the initial entity types do not match the expected type from the prediction, we use the category with the second highest probability, or the next highest probability until we receive a plausible entity pair. Naturally, the output category No_Relation is unaffected by these transformations.

## 3  Results

In this final section, we present the results of our experiments and the overall performance of our approach during the challenge. We discuss how our different modelling approaches performed better or worse, but also how our post-processing methods improved our accuracy, leading to the final, winning F1 score.

As mentioned in the methodology section, we explored three different approaches for training our model, during the development, but also the final phase. We then compared the performance of our approaches against each other, but also against the other competitor's baseline results in the final phase. The table below summarizes the F1 scores achieved by training our model using the different experimental approaches and language models. It highlights the superior performance of the first combination of language model and approach in terms of F1 score, confirming our earlier observations.

**TABLE 3: Performance by Model and Approach**

| Model | Approach | F1 Score |
| --- | --- | --- |
| Roberta-Large | Added marker before the entity | 0.726 |
| Bert-Base | Added marker before the entity | 0.697 |
| Roberta-Large | Wrapped marker around entity | 0.716 |
| Roberta-Large | Added Entity Pair before text | 0.637 |

After obtaining the predictions from our model using the first approach, we applied post-processing techniques to further refine the results, as described above. This involved implementing additional steps to remove false positives and enhance overall accuracy. In order to do that, we conducted an error analysis to gain insights into our model's limitations and areas for improvement. Through this analysis, we identified common patterns of misclassifications, false positives, and false negatives. For example, in cases where certain instances were classified as a relation that differed from the actual two entities (e.g., if the entities were "org:org" but the model returned "pers:org:founded_by"), we implemented a strategy to select the relation with the second highest probability, as the first classification was deemed incorrect. These findings lead to further refinements of our results, leading to the final 0.75 F1 scores.